%% file: main.tex
\documentclass[10pt,twocolumn,letterpaper]{article}

\usepackage{cvpr}      

\usepackage{multirow}
\usepackage{color}
\usepackage{stfloats}


\usepackage{array}
\usepackage{ragged2e}
\definecolor{cvprblue}{rgb}{0.21,0.49,0.74}
\usepackage[pagebackref,breaklinks,colorlinks,allcolors=cvprblue]{hyperref}

\def\confName{CVPR}
\def\confYear{2025}

\title{RoboGSim: A Real2Sim2Real Robotic Gaussian Splatting Simulator}

\author{Xinhai Li$^{1}$\thanks{\tt\small Equal contribution},\quad Jialin Li$^{2}$\footnotemark[1],\quad Ziheng Zhang$^{3}$\thanks{\tt\small Project leader},\quad Rui Zhang$^{4}$,\quad Fan Jia$^{3}$, \quad Tiancai Wang$^{3}$,\\ Haoqiang Fan$^{3}$, \quad Kuo-Kun Tseng$^{1}\thanks{\tt\small Corresponding authors}$,\quad  Ruiping Wang$^{2}$\footnotemark[3]\\$^1$Harbin Institute of Technology, Shenzhen\quad \\
$^2$Institute of Computing Technology, Chinese Academy of Sciences\\
$^3$MEGVII Technology\quad 
$^4$Zhejiang University
}

\begin{document}
\maketitle



\input{sec/0_abstract}    
\input{sec/1_intro}

\input{sec/2_related_work}
\input{sec/3_methods}
\input{sec/4_experiments}
\input{sec/5_conclusion}

{
    \small
    \bibliographystyle{ieeenat_fullname}
    \bibliography{main}
}

\clearpage
\end{document}

%% file: sec/0_abstract.tex
\begin{abstract}
Efficient acquisition of real-world embodied data has been increasingly critical. However, large-scale demonstrations captured by remote operation tend to take extremely high costs and fail to scale up the data size in an efficient manner. Sampling the episodes under a simulated environment is a promising way for large-scale collection while existing simulators fail to high-fidelity modeling on texture and physics.
To address these limitations, we introduce the RoboGSim, a real2sim2real robotic simulator, powered by 3D Gaussian Splatting and the physics engine. 
RoboGSim mainly includes four parts: Gaussian Reconstructor, Digital Twins Builder, Scene Composer, and Interactive Engine.
It can synthesize the simulated data with novel views, objects, trajectories, and scenes. RoboGSim also provides an online, reproducible, and safe evaluation for different manipulation policies. The real2sim and sim2real transfer experiments show a high consistency in the texture and physics. We compared the test results of RoboGSim data and real robot data on both RoboGSim and real robot platforms. The experimental results show that the RoboGSim data model can achieve zero-shot performance on the real robot, with results comparable to real robot data. Additionally, in experiments with novel perspectives and novel scenes, the RoboGSim data model performed even better on the real robot than the real robot data model. This not only helps reduce the sim2real gap but also addresses the limitations of real robot data collection, such as its single-source and high cost. We hope RoboGSim serves as a closed-loop simulator for fair comparison on policy learning. More information can be found on our project page 
\href{https://robogsim.github.io/}{https://robogsim.github.io/}.

\end{abstract}

%% file: sec/1_intro.tex
\section{Introduction}
\label{sec:intro}
Collecting large-scale manipulated data is of great importance for efficient policy learning. Some methods propose to capture the demonstrations as well as the actions through the remote operation~\cite{vr, aloha, mobilealoha}. While such operation relatively improves the collection efficiency, it tends to bring extremely large costs with the increasing data size. To solve this problem, some works~\cite{huber2024domain, tobin2017domain} attempt to generate the synthetic data under the simulated environment, which is further used to learn the manipulation policy. However, those Sim2Real approaches suffer from the large domain gap between simulated and real-world environments, making the learned policy invalid.

Recently, some works introduce the Real2Sim2Real (R2S2R) paradigm for robotic learning~\cite{nerf2real, lou2024robo}. The core insight is to perform realistic reconstruction via radiance field methods, such as NeRF~\cite{NeRF} and 3D Gaussian Splatting (3DGS)~\cite{3DGS}, and insert learned representations into the simulator. 
Among those methods, the typical approach, Robo-GS~\cite{lou2024robo}, presents a Real2Sim pipeline and introduces a hybrid representation to generate digital assets enabling high-fidelity simulation. However, it lacks the demonstration synthesis on novel scenes, views, and objects, as well as verification as policy learning data. Moreover, it fails to perform closed-loop evaluation for different policies due to the misalignment between the latent representation, simulation, and real-world spaces.

In this paper, we develop a Real2Sim2Real simulator, called RoboGSim, for both high-fidelity demonstration synthesis and physics-consistent closed-loop evaluation. It mainly includes four parts: Gaussian Reconstructor, Digital Twins Builder, Scene Composer and Interactive Engine. 
Given the multi-view RGB image sequences and MDH ~\cite{corke2007simple} parameters of the robotic arm, Gaussian Reconstructor is built upon 3DGS~\cite{mass3dgs} and reconstructs the scene and objects.
Then, the Digital Twins Builder performs the mesh reconstruction and creates a digital twin in Isaac Sim. In Digital Twins Builder, we propose the layout alignment module to align the space between the simulation, real-world, and GS representation.
After that, the Scene Composer combines the scene, robotic arm and objects in simulation, and renders the images from new perspective. Finally, in the Interactive Engine, RoboGSim works as the Synthesizer and Evaluator to performs the demonstration synthesis and closed-loop policy evaluation.

RoboGSim brings many advantages compared to existing (Real2)Sim2Real frameworks. It is the first neural simulator that unifies the demonstration synthesis and closed-loop evaluation. RoboGSim can generate realistic manipulated demonstrations with novel scenes, views, and objects for policy learning. It can also perform closed-loop evaluation for different policy networks, ensuring fair comparison under a realistic environment. In conclusion, our core contributions can be concluded as:
\begin{itemize}
\item[$\bullet$] \textbf{Realistic 3DGS-Based Simulator:} We develop a 3DGS-based simulator that reconstructs scenes and objects with realistic textures from multi-view RGB videos. RoboGSim is optimized for some challenging conditions like weak textures, low light, and reflective surfaces.
\item[$\bullet$] \textbf{Digital Twin System:} We introduce the layout alignment module in the system. With the layout-aligned Isaac Sim, RoboGSim maps the physical interactions between objects and robotic arms from Real2Sim spaces. 
\item[$\bullet$] \textbf{Synthesizer and Evaluator:} RoboGSim can synthesize the realistic manipulated demonstrations with novel scenes, views, and objects for policy learning. It can also work as the Evaluator to perform model evaluation in a physics-consistent manner.The experiment results show that our generated data can achieve the same performance as real robot data, which in a way solves the sim2real gap problem. At the same time, in experiments with novel scenes and novel perspectives, our generated data is more effective than real robot data, even with the 2D Aug method. Our evaluator also partially verifying the performance of the real robot model in a closed-loop.

\end{itemize}

%% file: sec/2_related_work.tex
\section{Related Work}
\label{sec:formatting}

\subsection{Sim2Real in Robotics}
The Real2Sim2Real approach fundamentally seeks to address the Sim2Real gap, which remains a persistent obstacle in the transformation from simulation to real world~\cite{20years,robosim2real}.
In order to bridge the Sim2Real gap as much as possible, many feature-rich simulators have emerged in recent years, including~\cite{isaacgym,isaacsim,pybullet,mujoco,sapien}. To this end, various datasets and benchmarks have also been proposed for effective policy learning~\cite{RoboHive,furniturebench,orbit,partmanip}.

Previous Sim2Real methods can be broadly classified into three categories: domain randomization, domain adaptation, and learning with disturbances~\cite{s2r_survey}. Domain randomization methods are designed to expand the operational envelope of a robot in a simulator by introducing randomness. The simulation environment should be capable of migration of the aforementioned capabilities in real-world settings~\cite{tobin2017domain,exarchos2021policy,andrychowicz2020learning,huber2024domain}. Domain adaptation approaches aim to unify the feature space of simulated and real environments, facilitating the training and migration within the unified feature space~\cite{bousmalis2017unsupervised,long2015learning,zheng2023gpdan}. The objective of learning methods introduce the disturbances into the simulated environment, in which the policy of robots is learned. It develops the capacity to operate effectively in the real world with noise and unpredictability~\cite{wang2020reinforcement,chi2024diffusionpolicy}.
\subsection{3D Gaussian Splatting in Robotics}
As a significant advancement in the field of 3D reconstruction, 3DGS~\cite{3DGS} represents the scene as a large set of explicit Gaussian points and combines it with efficient rasterization to achieve high-fidelity real-time rendering, extending the capabilities of NeRF~\cite{NeRF}.

More recently, a number of studies have explored the use of 3DGS to perform manipulation tasks within embodied simulators and the real world.
For example, ManiGaussian~\cite{ManiGaussian} introduces a dynamic GS framework alongside a Gaussian world model, which respectively represents Gaussian points implicitly and parameterizes them to model and predict future states and actions. Similarly, GaussianGrasper~\cite{GaussianGrasper} utilizes RGB-D images as inputs and embeds semantic and geometric features into 3DGS through feature distillation and geometric reconstruction, thereby enabling language-guided grasping operations.
To effectively transfer the knowledge learned in simulation to the real world and reduce the Sim2Real gap, recent works~\cite{lou2024robo,SplatSim,DifferentiableRobot} based on 3DGS have appeared. Among them, the most similar to ours are Robo-GS~\cite{lou2024robo} and SplatSim~\cite{SplatSim}. Robo-GS achieves manipulable robotic arm reconstruction by binding Gaussian points, grids, and pixels, with a primary focus on high-fidelity Real2Sim transfer; however, it provides limited discussion on the Sim2Real phase. SplatSim reconstructs both the robotic arm and objects in the scene and simultaneously verifies the feasibility of the method for Sim2Real tasks. However, it lacks discussions on generating digital twin assets of the objects, which are critical for achieving accurate manipulation.

%% file: sec/3_methods.tex
\newcommand{\project}[0]{\pi}
\newcommand{\meanW}[0]{\mu_{W}}
\newcommand{\meanC}[0]{\mu_{C}}
\newcommand{\meanI}[0]{\mu_{I}}
\newcommand{\covW}[0]{\Sigma_{W}}
\newcommand{\covI}[0]{\Sigma_{I}}
\newcommand{\cam}[0]{T}
\newcommand{\camCW}[0]{\cam_{CW}}
\newcommand{\camWC}[0]{\cam_{WC}}
\newcommand{\rotCW}[0]{R_{CW}}
\newcommand{\rotWC}[0]{R_{WC}}
\newcommand{\tauC}[0]{\tau}
\newcommand{\thetaC}[0]{\theta}
\newcommand{\Exp}[0]{\text{Exp}}
\newcommand{\Log}[0]{\text{Log}}
\newcommand{\loss}[0]{\mathcal{L}}
\newcommand{\meddepth}[0]{\tilde{\mathcal{D}}}
\newcommand{\gaussians}[0]{\mathcal{G}}
\newcommand{\window}[0]{\mathcal{W}}
\newcommand{\kfwindow}[0]{\mathcal{W}_k}
\newcommand{\randwindow}[0]{\mathcal{W}_r}
\newcommand{\pd}[2]{\frac{\partial {#1} }{\partial {#2} }}
\newcommand{\mpd}[2]{\frac{\mathcal{D} {#1}}{\mathcal{D} {#2}}}
\newcommand{\se}[1]{\mathfrak{se}(#1)}
\newcommand{\SE}[1]{SE(#1)}
\newcommand{\identity}[0]{I}
\newcommand{\vectorize}[1]{vec(#1)}
\newcommand{\kron}[0]{\otimes}
\newcommand{\RR}{\mathbb{R}}
\newcommand{\gtimage}[0]{\bar{I}}
\newcommand{\gtdepth}[0]{\bar{D}}
\newcommand{\matJ}[0]{\mathbf{J}}
\newcommand{\matW}[0]{\mathbf{W}}
\newcommand{\vecs}[0]{\mathbf{s}}

\begin{figure*}[htbp]
\includegraphics[width=\linewidth]{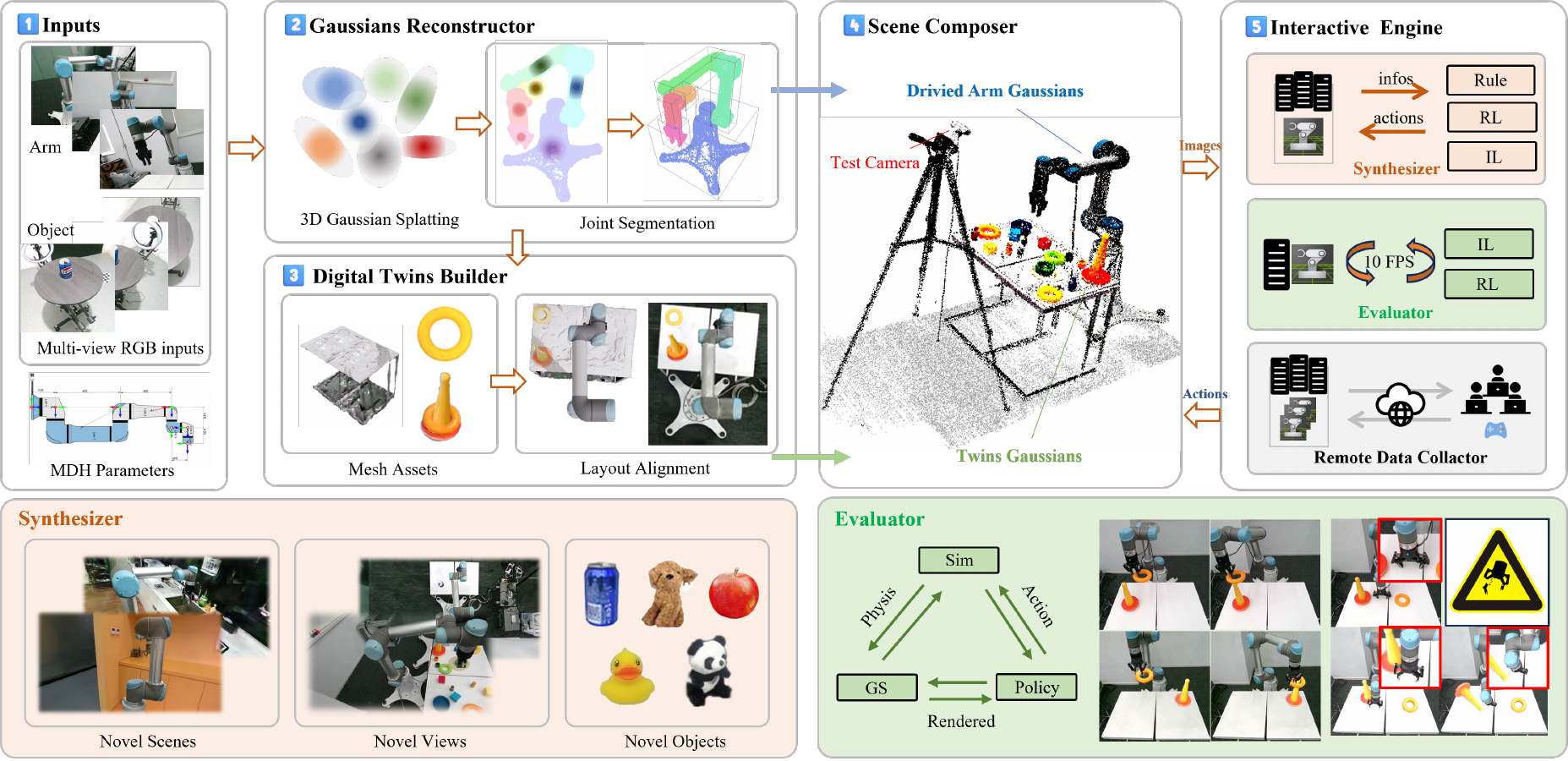}
            \captionof{figure}{\label{fig:pipeline} \textbf{Overview of the RoboGSim Pipeline:} (1) Inputs: multi-view RGB image sequences and MDH parameters of the robotic arm. (2) Gaussian Reconstructor: reconstruct the scene and objects using 3DGS, segment the robotic arm and build an MDH kinematic drive graph structure for accurate arm motion modeling. (3) Digital Twins Builder: perform mesh reconstruction of both the scene and objects, then create a digital twin in Isaac Sim, ensuring high fidelity in simulation. (4) Scene Composer: combine the robotic arm and objects in the simulation, identify optimal test viewpoints using tracking, and render images from new perspectives. (5) Interactive Engine: (i) The synthesized images with novel scenes/views/objects are used for policy learning. (ii) Policy networks can be evaluated in a close-loop manner. (iii) The embodied data can be collected by the VR/Xbox equipment. 
            }
\vspace{-1em}
\end{figure*}    

\section{Methods}
\subsection{Overall Architecture}
As shown in Fig.~\ref{fig:pipeline}, RoboGSim mainly includes four parts: Gaussian Reconstructor, Digital Twins Builder, Scene Composer, and Interactive Engine. Given multi-view images and MDH parameters of the robotic arm, Gaussian Reconstructor (Sec.~\ref{Gaussian_Reconstructor}) reconstructs scenes and objects using 3DGS, segments the robotic arm, and builds an MDH kinematic drive graph structure to enable accurate motion modeling of the robotic arm. Digital Twin Builder (Sec.~\ref{Digital_Twins_Builder}) involves mesh reconstruction of the scene and objects. Through layout alignment, the asset data flow can be interconnected, facilitating the subsequent evaluation in Interactive Engine. Scene Composer (Sec.~\ref{Scene_Composer}) achieves the synthesis of novel objects, scenes, and views. Interactive Engine (Sec.~\ref{Interactive_Engine}) synthesizes novel view/scene/object images for policy learning. It can also evaluate the policy networks in a closed-loop manner. Moreover, we can collect the manipulated data in simulation using VR/Xbox equipment of the real world.

\subsection{Gaussian Reconstructor}

\label{Gaussian_Reconstructor}

We employ the 3DGS method to reconstruct static scenes, followed by point cloud segmentation of the robotic arm’s joints. Subsequently, we utilize the MDH dynamic model to control the point clouds corresponding to each joint, facilitating the dynamic rendering of the robotic arm.

3D Gaussian Splatting (3DGS)~\cite{3DGS} employs a set of multi-view images as input to achieve high-fidelity scene reconstruction. 3DGS represents the scene as a set of Gaussians and utilizes a differentiable rasterization rendering method to enable real-time rendering. 

Specifically, for a scene $\mathcal{G}=\left \{ \mathit{g}_i\right \}_{i=1}^{N}$ represented by $N$ Gaussians, each Gaussian can be represented as $g_i = {(\mu_i, \Sigma_i, o_i, c}_i)$. Here, $\mu \in \mathbb{R}^3$, $\Sigma\in\mathbb{R}^{3\times3}$, $o \in \mathbb{R}$ and $c \in SH(4)$ denote the mean, covariance matrix, opacity and color factor, represented by spherical harmonic coefficients, respectively.

During the rendering process, the final color value $C$ of the pixel can be obtained through a rendering method, similar to alpha-blending~\cite{3DGS}. It utilizes a sequence of $N$ ordered Gaussians that overlap with the pixel. Such process can be expressed as follows:
\begin{equation}
  C=\sum_{i\in N}c _i\alpha_i \prod_{j=1}^{i-1}(1-\alpha_j)
  \label{elpha_blending}
\end{equation}
\begin{equation}
\alpha_i = o_i\cdot \exp(\frac{1}{2}\delta_i^\top \Sigma_{2D}^{-1}\delta_i)
\end{equation}
where $\alpha_i$ is the opacity of the $i$-th Gaussian. $\delta_i \in \mathbb{R}^2$ denotes the offset between 2D Gaussian center and current pixel. $\Sigma_{2D} \in \mathbb{R}^{2\times2}$ represents the 2D covariance matrix.

 Modified Denavit-Hartenberg (MDH)~\cite{corke2007simple} convention is a parameterized model to describe the kinematic chain of a manipulator. Each joint and link in the kinematic chain is characterized by a set of parameters. In MDH, a transformation matrix can be constructed for each link, achieving an accurate representation of the manipulator's pose at each stage of motion.
 Let ${x_i, y_i, z_i}$ denote the coordinates of the origin for the $i$-th joint. For a manipulator, the $i$-th joint configuration can be represented as:
\begin{equation}
\Theta = \{\beta_i, a_i, d_i, \theta_i\}
\end{equation}
where $\beta_i$ represents the twist angle, which is the rotation around the $x$-axis from the $(i-1)$-th joint to the $i$-th joint. $a_i$ denotes the link length, measuring the distance along the $x$-axis from $z_{i-1}$ to $z_i$. $d_i$ is the link offset, indicating the displacement along the $z$-axis from $x_{i-1}$ to $x_i$. $\theta_i$ represents the joint angle, rotation around the $z$-axis from $x_{i-1}$ to $x_i$.

The transformation matrix for each link $T_i$, using MDH parameters, can be written as:
\begin{equation}
\setlength{\arraycolsep}{2pt}
T_i \!=\! \begin{bmatrix}
\cos \theta_i & -\sin \theta_i \cos \beta_i & \sin \theta_i \sin \beta_i & a_i \cos \theta_i \\
\sin \theta_i & \cos \theta_i \cos \beta_i & -\cos \theta_i \sin \beta_i & a_i \sin \theta_i \\
0 & \sin \beta_i & \cos \beta_i & d_i \\
0 & 0 & 0 & 1
\end{bmatrix}
\label{eq:T_i}
\end{equation}
By sequentially multiplying these transformation matrices, we can obtain the final transformation matrix from the base to the end effector.
We segment each joint and then treat all Gaussian points within a joint as a point mass. We further move all Gaussian points within a joint according to $T_i$, achieving kinematic-driven Control of the Gaussian points.

\subsection{Digital Twins Builder}
\label{Digital_Twins_Builder}

Digital twins should not only map real-world assets but also involve coordinate alignment. Through Real2Sim layout alignment and Sim2GS sparse keypoint alignment, we can digitize the real world, enabling the flow of digital assets between the real, simulated, and GS representation. This facilitates the conversion of digital assets in all directions, achieving comprehensive asset flooding.

\noindent \textbf{3D Assets Generation}: We employ two methods to generate 3D object assets. For real-world objects, we capture high-quality multi-view images of the objects using a turntable and extract matching features with GIM~\cite{gim} to address issues such as lack of texture and reflections. We then integrate the COLMAP pipeline~\cite{colmap} to obtain the initial SFM point cloud, which is subsequently used for reconstruction by 3DGS. Moreover, for novel objects available on the web, we initially employ generative 3D reconstruction methods \cite{wonder3d,li2023gaussiandiffusion} to procure 3D gaussians and textured meshes of the objects. Subsequently, we utilize the method in GaussianEditor~\cite{Chen_2024_CVPR} that applies the diffusion model~\cite{stablediffusion} to facilitate object reconstruction in 3DGS.

\noindent \textbf{Layout Alignment}: As shown in Fig.~\ref{fig:pipeline}, since we follow the local coordinate system of the robotic arm, the world coordinates and Isaac Sim are axis-aligned. We first measure the real-world scene to align the size of the imported table scene in Isaac Sim. In the GS scene, a downward-facing camera is placed 1.6 meters above the base joint to render a segmentation map. For coordinate alignment, we place a downward-facing camera 1.6 meters above the base joint in Isaac Sim. By comparing the rendered scene from the BEV, front and side view segmentation, with the views from Isaac Sim, we adjust the shift to achieve layout alignment.

\noindent \textbf{Sim2GS Alignment}: Given the MDH-based transformation matrices $ T_i^{gs}$ and simulated transformation matrices $ T_i^{sim}$, there exists a transformation matrix $T_{gs \ (i)}^{sim}$ such that:
\begin{equation}
T_{gs \ (i)}^{sim}= T_i^{sim} \cdot T_{gs}^{i}
\end{equation}
To compute the average transformation matrix $T_{gs}^{sim}$, we use the weighted sum and apply normalization:
\begin{equation}
  T_{gs}^{sim}= \frac{\sum_{i=1}^{6} w_i \cdot T_{gs \ i}^{sim}}{\left\| \sum_{i=1}^{6} w_i \cdot T_{gs \ i}^{sim} \right\|}
\end{equation}
where $w_i$ is the weight of each joint. 

For the target object $T_{obj}^{sim}$in  Isaac Sim, we can transform it into the GS coordinate system using the following formula:
\begin{equation}
T_{obj}^{gs}=  T_{sim }^{gs} \cdot  T_{obj}^{sim}
\label{eq6}
\end{equation}

\noindent\textbf{Camera Localization}: To transform the real-world coordinate system into the GS coordinate, we apply the localization approach from GS-SLAM~\cite{GS_SLAM}. For a pre-trained GS model, $\mathcal{G}=\left \{ \mathit{g}_i\right \}_{i=1}^{N}$ , we froze the attributes of 3DGS and optimize the external camera parameters $\cam_C^W$. 

In camera localization, only the current camera pose is optimized without updates to the map representation. For monocular cases, we minimize the following photometric residual:
\begin{equation}
    \mathcal{L}_{pho} = \left\| I(\gaussians, \cam_C^W) - \gtimage \right\|_1~,
    \label{eqn:photometric}
\end{equation}
where $I(\gaussians, \cam_C^W)$ represents rendering Gaussians $\gaussians$ from $\cam_C^W$, and $\gtimage$ is the observed image.

\subsection{Scene Composer}
\label{Scene_Composer}

\noindent\textbf{Scene Editing}:
To merge the point cloud into the robotic arm scene, the transformation $T[R|t]$ of the marked point is first calculated. Then the coordinates of the point cloud in the new scene are projected into the arm coordinate based on the transformation. Expanding the 3D Covariance $\Sigma$ in 3DGS into scale $s$ and rotation quaternion $q$ by:
\begin{equation}
    \Sigma = qss^Tq^T
  \label{eq:7}
\end{equation}

The ratio $r$ of the transformation can be isolated and extracted as an independent component:
\begin{equation}
   r = \sqrt{(R R^T)_{(0,0)}}
\end{equation}
 we can further use it to normalize the rotation matrix $R$:
\begin{equation}
   R_{\text{norm}} = \frac{R}{r}
\end{equation}
The scale attribute \( s \) of the Gaussian points is adjusted:
\begin{equation}
   \text{s} = \text{s} + \log(r) \quad 
\end{equation}
Apply the Transformation $T$ to Gaussian point coordinates
\begin{equation}
    \mu^\prime = R\mu+t
  \label{trans_mu}
\end{equation}
\begin{equation}
    \Sigma^\prime = R_{norm}\Sigma R_{norm}^\top
  \label{trans_sigma}
\end{equation}

\noindent\textbf{Object Editing}:
The transformation here can extend the transformation from the scene editing mentioned above. However, the difference is that the target object's coordinate center is given by Eq.~\ref{eq6}. The coordinate transformation for its Gaussian points can be represented:
\begin{equation}
    \mu^\prime = R(\mu-\mu_0)+\mu_0+t
  \label{trans_mu_local}
\end{equation}

\subsection{Interactive Engine}
\label{Interactive_Engine}

Our interactive engine can work as: Synthesizer and Evaluator. As Synthesizer, it produces large volumes of data with low-cost for downstream policy learning. As Evaluator, it can perform safe, real-time, and reproducible evaluation.

\noindent\textbf{Synthesizer}: We use the engine to generate numerous training trajectories, including robotic arm movements and target object trajectories. These trajectories drive the GS to generate massive and photorealistic simulated datasets for policy learning. This diverse data includes novel view renderings, scene combinations, and object replacements.

\noindent\textbf{Evaluator}: For trained models, testing directly on physical devices may pose safety risks or incur high costs for reproduction. Therefore, we convert the predicted trajectories into GS-rendered results to efficiently and rapidly evaluate the model's prediction quality. Specifically, the Isaac Sim~\cite{isaacsim} outputs an initial state of the target object and robotic arm, and GS renders according to the status. The rendered images are then fed to the policy to predict the next frame's action. The predicted action is passed to the simulation for kinematic inverse parsing, collision detection, and other physical interactions. Then, Isaac sim sends the parsed six-axis relative pose to the GS renderer, which then sends the rendered result as feedback to the policy network. This serves as visual feedback for predicting the next action, and the process iterates until the task is finished.

%% file: sec/4_experiments.tex
\section{Experiments}

Since there is no benchmarks available for Real2Sim2Real, we construct the following four groups of proxy experiments to comprehensively evaluate the performance of RoboGSim under simulation and real-world. We use UR5 robot arm for all experiments. 
The robot arm rendering is partially built upon the codebase of Robo-GS~\cite{lou2024robo}.

\noindent \textbf{Real2Sim Novel Pose Synthesis} verifies whether the robot arm pose captured in the real world can be effectively utilized to achieve precise control in the simulator.

\noindent \textbf{Sim2Real Trajectory Replay}
checks whether the trajectories collected in the simulator can be accurately reproduced by the real-world robot arm.

\noindent \textbf{RoboGSim as Synthesizer} demonstrates the ability of RoboGSim to generate high-fidelity demonstrations with novel scenes, views, and objects, aligning with real world.

\noindent \textbf{RoboGSim as Evaluator} shows that RoboGSim can effectively perform closed-loop evaluation for policy networks.

\begin{table*}[ht]
\centering
\begin{tabular}{|c|c|c|c|c|c|c|}
\hline
 & \multicolumn{3}{c|}{Grasp Suc.} & \multicolumn{3}{c|}{Place Suc.} \\
\hline
 & TV & NV(MD) & NS & TV & NV & NS \\
\hline
Real & 100\% &  30\%  & 40\% & 90\% & 0\% & 20\% \\
Real+2D AUG & 80\% &  100\%  & 60\% & 80\% & 0\% & 60\% \\
RoboGSim & 100\% &  70\%  & 100\% & 90\% & 0\% & 90\% \\
\hline
\end{tabular}
\caption{ Performance on ring-toss task in real world. TV denotes test view, NV is novel view and NS is the novel scene. MD means minor deviation.
}
\label{grasp_place}
\end{table*}

\begin{table}[ht]
\centering
\begin{tabular}{|>{\raggedright}m{4cm}|c|c|}
\hline
\textbf{Method} & \textbf{Grasp Suc.} & \textbf{Place Suc.} \\
\hline
Real-to-Real & 100\% & 90\% \\
Real-to-RoboGSim & 100\% & 30\% \\
Sim-to-Real & 80\% & 0\% \\
RoboGSim-to-Real & 100\% & 90\% \\
\hline
\end{tabular}
\caption{Cross validation between real world and simulation. “Sim-to-Real” means that testing the VLA model trained with the Isaac sim data on the real world robot arm.}
\end{table}

\subsection{Real2Sim Novel Pose Synthesis}

\begin{figure*}[!htbp]
\centering
\includegraphics[scale=0.56]{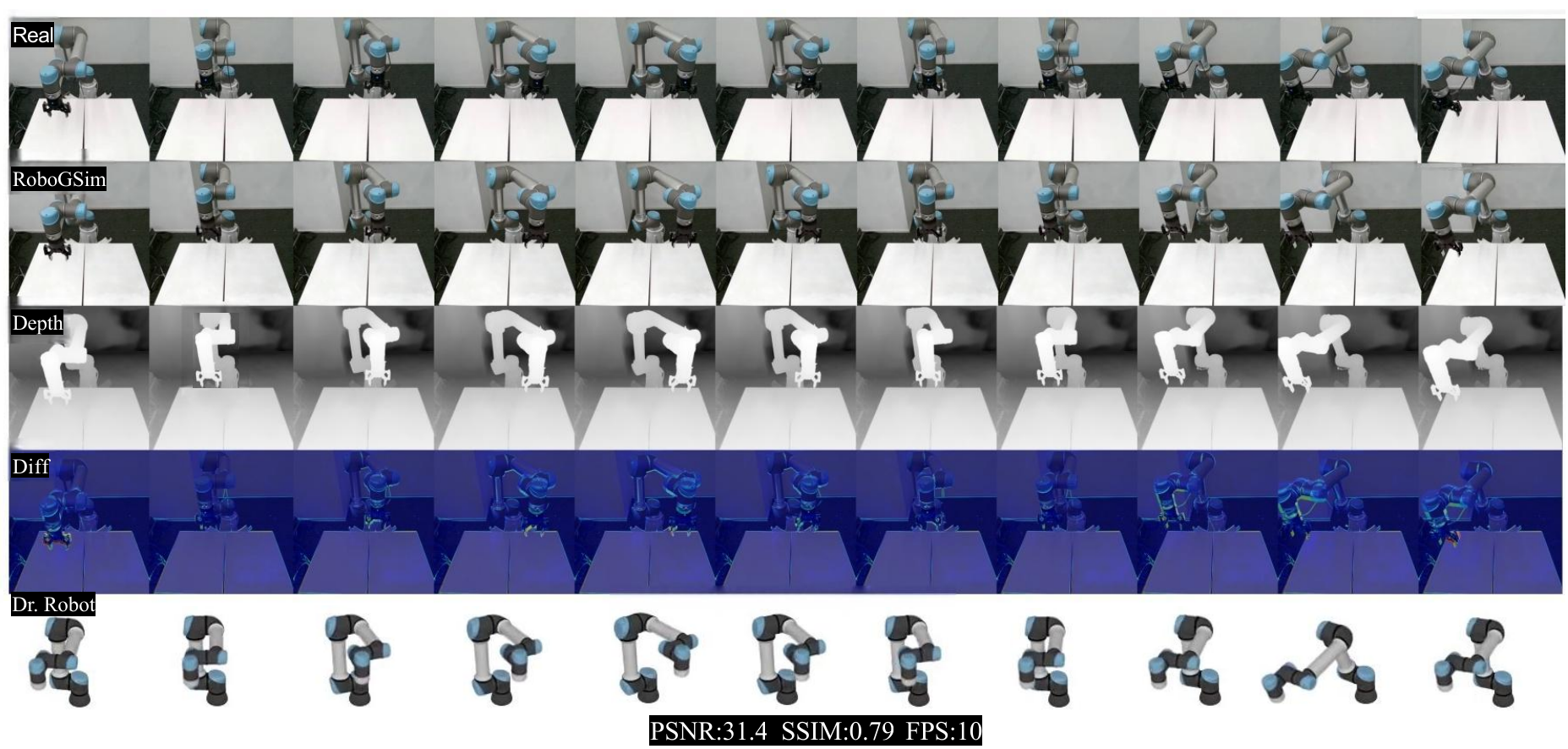}
\captionof{figure}{\label{fig:exp1} \textbf{Real2Sim Novel Pose Synthesis:}
 "Real" represents the capture of the real robotic arm from a new viewpoint. "RoboGSim" shows the rendering of the novel pose from the new viewpoint driven by the real recorded trajectory. "Depth" shows the rendering depth by GS. "Diff" is the difference calculated between the Real and the rendered RGB images. We compute the pixel distance of the same point between the Real and RoboGSim, which is 7.37. }
\label{exp1}
\end{figure*}

The objective of the novel pose synthesis is to validate the performance of Real2Sim reconstruction, with a particular focus on the accuracy of the robotic arm's movements and the fidelity of the image texture. The static scene is reconstructed using the initial pose of the robotic arm from the first frame of GT. The trajectory collected from the real robotic arm is used as the driving force, and we employ the kinematic control for novel pose rendering. 
As shown in Fig.~\ref{exp1}, the results demonstrate that our reconstruction accurately captures both the texture and the physical dynamics of the robotic arm, highlighting the fidelity achieved by RoboGSim. To compare with the video sequence driven by the real robot under the new viewpoint, RoboGSim achieves a 31.3 PSNR and 0.79 SSIM rendering result, while ensuring real-time rendering with 10 FPS.

\begin{figure*}[!htb]
\centering
\includegraphics[scale=0.51]{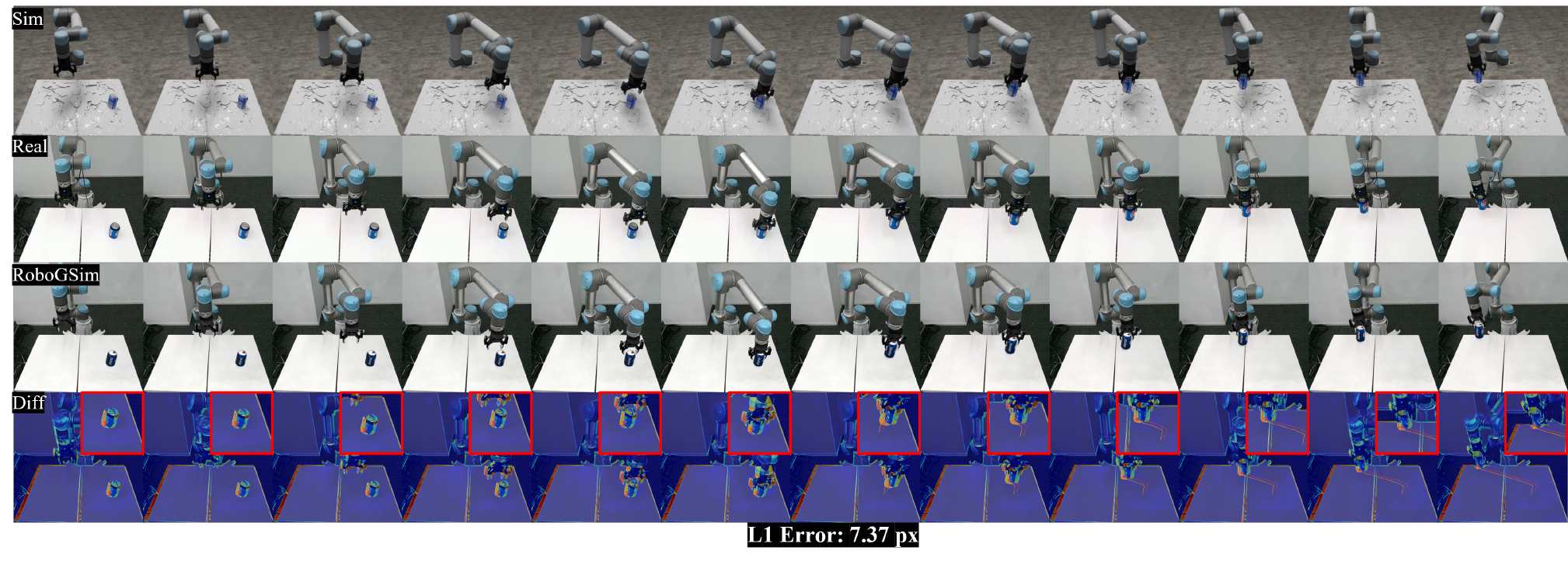}
\captionof{figure}{\label{fig:exp2} \textbf{Sim2Real Trajectory Replay:}
The "Sim" row displays the video sequence collected from Isaac Sim. "Real" represents the demonstration driven by the trajectory in simulation. "RoboGSim" is the GS rendering result driven by the same trajectory. "Diff" indicates the differences between Real and the rendered results. }
\label{exp2}
\end{figure*}
\begin{figure*}[!htb]
\centering
\includegraphics[scale=0.55]{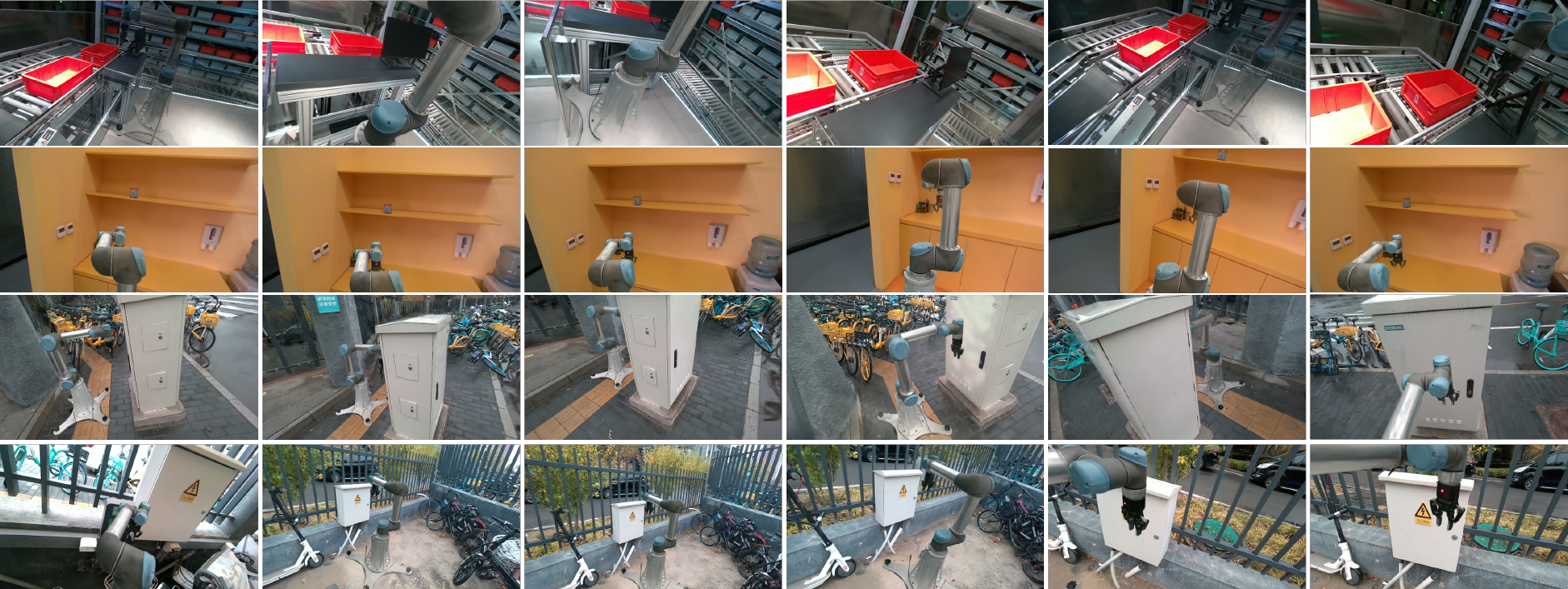}
\captionof{figure}{\label{fig:novel_scene} \textbf{Novel Scene Synthesis:} We show the results of the physical migration of the robot arm to new scenes, including a factory, a shelf, and two outdoor environments. The high-fidelity multi-view renderings demonstrate that RoboGSim enables the robot arm to operate seamlessly across diverse scenes.}
\label{novel_scene}
\end{figure*}
\subsection{Sim2Real Trajectory Replay}

To verify whether the trajectories from Issac Sim can perfectly align with the real machine and RoboGSim, we designed an experiment where the trajectory is collected using Issac Sim, and then the trajectory is used to drive GS to render a Coke-grasping scene, while the same trajectory is used to drive the real machine to grasp a Coke can. 
As shown in Fig.~\ref{exp2}, the comparison reveals a strong alignment between the simulated policy and the actual physical behavior of the robotic arm, highlighting the effectiveness of the Sim2Real transfer in our system. These results suggest that our simulation can reliably model real-world dynamics, facilitating successful policy transfer from simulation to the real world.

\subsection{RoboGSim as Synthesizer}

\begin{figure*}[!htb]
\centering
\includegraphics[scale=0.64]{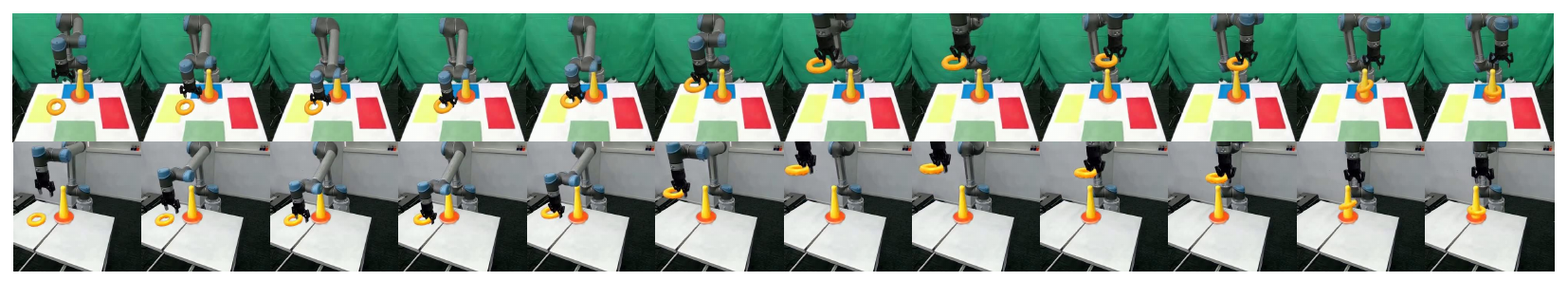}
\captionof{figure}{\label{fig:exp431}
\textbf{RoboGSim as Synthesizer: }
Rendering of the same training set in novel view and novel scenes. }
\label{exp431}
\end{figure*}

\begin{figure*}[!htb]
\centering
\includegraphics[scale=0.59]{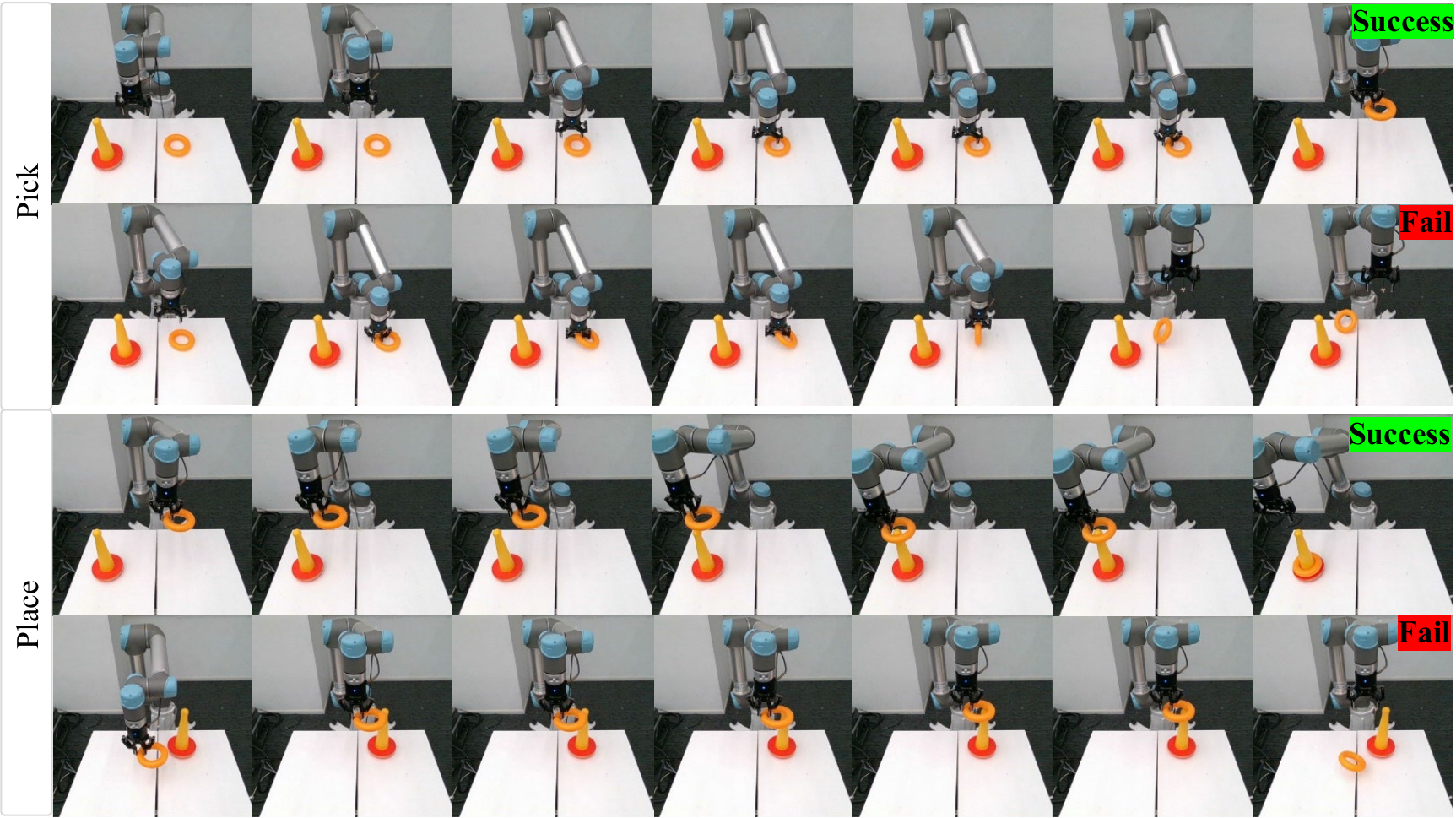}
\captionof{figure}{\label{fig:exp43}
\textbf{RoboGSim as Synthesizer:} The first two rows show real robot videos captured from the test viewpoint, illustrating successful and failed cases of the VLA model on the Pick task. The last two rows display real robot videos captured from the test viewpoint, showing successful and failed cases of the VLA model on the Place task.}
\label{exp43}

\end{figure*}
\begin{figure*}[!htb]
\centering
\includegraphics[scale=0.55]{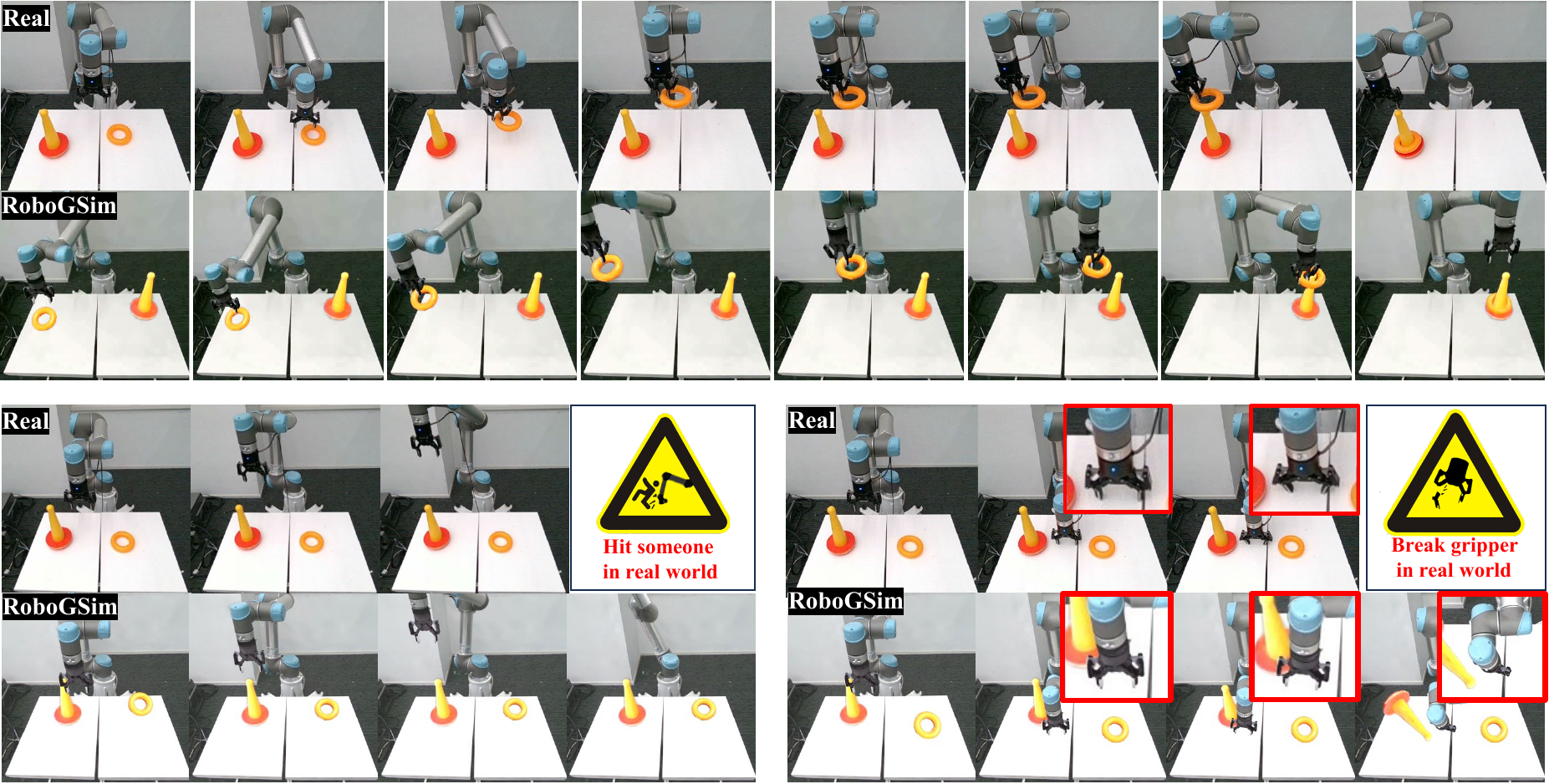}
\captionof{figure}{\label{fig:exp44} 
\textbf{RoboGSim as Evaluator}: The first two rows, labeled "Real" and "RoboGSim", show the footage captured from the real robot and RoboGSim, respectively. They are both driven by the trajectory generated by the same VLA network. In the third row, the left side shows the real-world inference where the robot arm exceeds its operational limits, resulting in a manual shutdown. The right side shows an instance where a wrong decision from the VLA network, causes the robotic arm to collide with the table. The fourth row presents the simulation results from RoboGSim, which can avoid dangerous collisions.}
\label{exp44}
\vspace{-1.5em}
\end{figure*}

\begin{figure*}[!htb]
\centering
\includegraphics[scale=0.62]{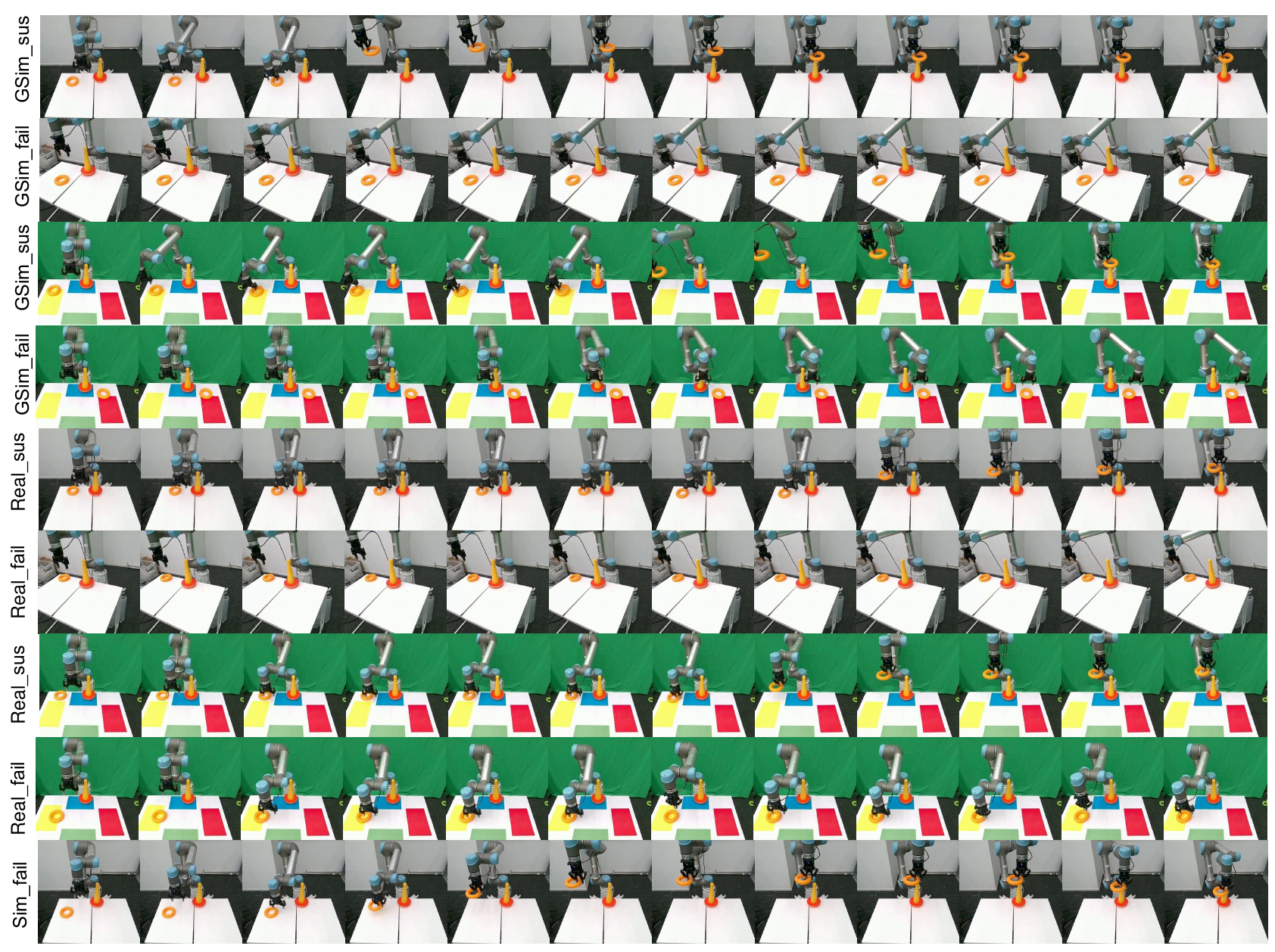}
\captionof{figure}{\label{fig:exp44} 
\textbf{Exp on  Real robot}: The first row, \texttt{GSim\_sus}, represents successful cases of RoboGSim data under the test view.  
The second row, \texttt{GSim\_fail}, represents failure cases of RoboGSim data under the novel view.  
The third row, \texttt{GSim\_sus}, represents successful cases of RoboGSim data under the novel scene.  
The fourth row, \texttt{GSim\_fail}, represents failure cases of RoboGSim data under the novel scene.  
The fifth row, \texttt{Real\_sus}, represents successful cases of real-world data under the test view.  
The sixth row, \texttt{Real\_fail}, represents successful cases of real-world data under the novel view.  
The seventh row, \texttt{Real\_sus}, represents successful cases of real-world data under the novel scene.  
The eighth row, \texttt{Real\_fail}, represents failure cases of real-world data under the novel scene.  
The ninth row, \texttt{Sim\_fail}, represents failure cases of Isaac Sim data under the test view.
}
\label{exp44}
\end{figure*}

In this section, we use the vision-language-action (VLA) model to validate the effectiveness of synthetic data by RoboGSim. We use the LLAMA3-8B~\cite{llama3} as the LLM and CLIP~\cite{CLIP} as the vision encoder. Two-layer MLP is used as the projection network. The VLA model is trained on 8xA100 (80GB) for 1 epoch. The training process is divided into three stages: (1) Pre-training with only the connector enabled, using the LAION-558K dataset. (2) Training with LLM unfrozen using the LLaVA665K dataset. (3) Supervised Finetuning (SFT) with robotic image-action data and the CLIP weight is frozen.

By using the real machine distribution to guide the RoboGSim distribution, we aim to improve the model's success rate. We perform the experiments on a challenging ring-toss task (see Fig.~\ref{exp43}), which is divided into two subtasks: picking up the ring and tossing it onto the target. The accuracy requirement for the Z-axis when picking up the ring is within 5mm. For real data, 1,000 samples are collected manually. For a fair comparison, we used 1,000 synthetic samples generated by RoboGSim. During testing, each model was tested 10 times, with three attempts allowed per trial for grasping. If all three attempts failed, the trial was marked as unsuccessful.

As shown in Tab.~\ref{grasp_place}, We compared three models: one trained with real machine data, one trained with real machine data plus 2D AUG, and one trained with RoboGSim. The comparison was made in terms of test view, novel view, and novel scene. The results show that in the test view, the generated data from RoboGSim can achieve zero-shot capability, with performance comparable to the real machine data (both at 90\%). In the novel scene case, RoboGSim performed much better than real machine data, reaching 90\% compared to the 60\% of real machine data. In the novel view experiment, RoboGSim also had less bias compared to the real machine data model. We also compared the effect of adding 2D AUG to the real machine data. After adding AUG, the performance in the test view dropped (90\% $\rightarrow$ 80\%), but in the novel scene, it improved (40\% $\rightarrow$ 60\%). However, in the novel view, the bias increased. Pure 2D AUG lacks spatial awareness, and its performance is far worse than that of RoboGSim, which has spatial intelligence.
a
It should be noted that manual collection takes a total of 40 hours while RoboGSim only requires 4 hours for synthesis. It is promising to further scale up the data size of synthesis for further performance improvements. Fig.~\ref{exp43} shows the visualization of some success and failure cases. Moreover, we also illustrate some more qualitative analysis for novel scene synthesis. As shown in Fig.~\ref{fig:novel_scene}, we display the results of the physical migration of the UR5 robot arm to new scenes, including a factory, a shelf, and two outdoor environments. The high-fidelity multi-view renderings demonstrate that RoboGSim enables the robot arm to operate
seamlessly across diverse scenes.



\begin{table}[ht]
\centering
\begin{tabular}{|c|c|c|}
\hline
\textbf{Method} & \textbf{$L_1$ ↓} & \textbf{PSNR ↑} \\
\hline
3DGS & 0.01381 & 34.19939 \\
2DGS & 0.01798 & 32.36417 \\
PGSR & 0.01925 & 31.72386 \\
MipNeRF360 & 0.02618 & 23.51348
 \\
\hline
\end{tabular}
\caption{ 3DGS achieves better static reconstruction than 2DGS}
\end{table}

\subsection{RoboGSim as Evaluator}

Realistic closed-loop evaluation is crucial for validation and comparison of policy networks.
In this section, we mainly explore the effectiveness of using RoboGSim as an Evaluator. It aims to show its high consistency with real-world inference.
Given the well-trained VLA model, we deploy it for both real-world robots and RoboGSim simulation. 
As shown in Fig.~\ref{exp44}, our closed-loop simulator RoboGSim can reproduce results similar to those from the real world. For similar bad cases, our RoboGSim can avoid the issues existing in the real world, like violations and collisions.
Therefore, our evaluator provides a fair, safe, and efficient evaluation platform for policy. 

%% file: sec/5_conclusion.tex
\section{Conclusion and Discussion}

In this paper, we built a Real2Sim2Real simulator, based on 3DGS. We also introduce the digital twin system with spatial alignment to enables 3D assert flow. With novel viewpoint, object, trajectory and scene, our RoboGSim engine can generate high-fidelity synthesized data. Additionally, due to our precise spatial alignment, RoboGSim can serve as evaluator that allows real-time online policy evaluation. Despite its great progress, the current version of RoboGSim has several limitations. It can only simulate rigid objects and the lighting for synthesized objects is not yet fully unified with the robotic arm. Moreover, generating geometrically consistent object meshes remains challenging, which is often key to completing complex manipulation tasks. In the near future, we will explore more advanced mesh extraction methods, further expand the task categories and establish the benchmarks to comprehensively evaluate the performance across diverse scenarios.

\section{Acknowledgements}
The work was supported by the National Science and Technology Major Project of China (2023ZD0121300).

%% file: main.bbl
\begin{thebibliography}{43}
\providecommand{\natexlab}[1]{#1}
\providecommand{\url}[1]{\texttt{#1}}
\expandafter\ifx\csname urlstyle\endcsname\relax
  \providecommand{\doi}[1]{doi: #1}\else
  \providecommand{\doi}{doi: \begingroup \urlstyle{rm}\Url}\fi

\bibitem[Andrychowicz et~al.(2020)Andrychowicz, Baker, Chociej, Jozefowicz, McGrew, Pachocki, Petron, Plappert, Powell, Ray, et~al.]{andrychowicz2020learning}
OpenAI:~Marcin Andrychowicz, Bowen Baker, Maciek Chociej, Rafal Jozefowicz, Bob McGrew, Jakub Pachocki, Arthur Petron, Matthias Plappert, Glenn Powell, Alex Ray, et~al.
\newblock Learning dexterous in-hand manipulation.
\newblock \emph{The International Journal of Robotics Research}, 39\penalty0 (1):\penalty0 3--20, 2020.

\bibitem[Bousmalis et~al.(2017)Bousmalis, Silberman, Dohan, Erhan, and Krishnan]{bousmalis2017unsupervised}
Konstantinos Bousmalis, Nathan Silberman, David Dohan, Dumitru Erhan, and Dilip Krishnan.
\newblock Unsupervised pixel-level domain adaptation with generative adversarial networks.
\newblock In \emph{Proceedings of the IEEE conference on computer vision and pattern recognition}, pages 3722--3731, 2017.

\bibitem[Byravan et~al.(2023)Byravan, Humplik, Hasenclever, Brussee, Nori, Haarnoja, Moran, Bohez, Sadeghi, Vujatovic, and Heess]{nerf2real}
Arunkumar Byravan, Jan Humplik, Leonard Hasenclever, Arthur Brussee, Francesco Nori, Tuomas Haarnoja, Ben Moran, Steven Bohez, Fereshteh Sadeghi, Bojan Vujatovic, and Nicolas Heess.
\newblock Nerf2real: Sim2real transfer of vision-guided bipedal motion skills using neural radiance fields.
\newblock In \emph{2023 IEEE International Conference on Robotics and Automation (ICRA)}, pages 9362--9369, 2023.

\bibitem[Chen et~al.(2024)Chen, Chen, Zhang, Wang, Yang, Wang, Cai, Yang, Liu, and Lin]{Chen_2024_CVPR}
Yiwen Chen, Zilong Chen, Chi Zhang, Feng Wang, Xiaofeng Yang, Yikai Wang, Zhongang Cai, Lei Yang, Huaping Liu, and Guosheng Lin.
\newblock Gaussianeditor: Swift and controllable 3d editing with gaussian splatting.
\newblock In \emph{Proceedings of the IEEE/CVF Conference on Computer Vision and Pattern Recognition (CVPR)}, pages 21476--21485, 2024.

\bibitem[Chi et~al.(2024)Chi, Xu, Feng, Cousineau, Du, Burchfiel, Tedrake, and Song]{chi2024diffusionpolicy}
Cheng Chi, Zhenjia Xu, Siyuan Feng, Eric Cousineau, Yilun Du, Benjamin Burchfiel, Russ Tedrake, and Shuran Song.
\newblock Diffusion policy: Visuomotor policy learning via action diffusion.
\newblock \emph{The International Journal of Robotics Research}, 2024.

\bibitem[Corke(2007)]{corke2007simple}
Peter~I Corke.
\newblock A simple and systematic approach to assigning denavit--hartenberg parameters.
\newblock \emph{IEEE transactions on robotics}, 23\penalty0 (3):\penalty0 590--594, 2007.

\bibitem[Coumans and Bai(2016--2021)]{pybullet}
Erwin Coumans and Yunfei Bai.
\newblock Pybullet, a python module for physics simulation for games, robotics and machine learning.
\newblock \url{http://pybullet.org}, 2016--2021.

\bibitem[Dimitropoulos et~al.(2022)Dimitropoulos, Hatzilygeroudis, and Chatzilygeroudis]{robosim2real}
Konstantinos Dimitropoulos, Ioannis Hatzilygeroudis, and Konstantinos Chatzilygeroudis.
\newblock A brief survey of sim2real methods for robot learning.
\newblock In \emph{International Conference on Robotics in Alpe-Adria Danube Region}, pages 133--140. Springer, 2022.

\bibitem[Dubey et~al.(2024)Dubey, Jauhri, Pandey, Kadian, Al-Dahle, Letman, Mathur, Schelten, Yang, Fan, et~al.]{llama3}
Abhimanyu Dubey, Abhinav Jauhri, Abhinav Pandey, Abhishek Kadian, Ahmad Al-Dahle, Aiesha Letman, Akhil Mathur, Alan Schelten, Amy Yang, Angela Fan, et~al.
\newblock The llama 3 herd of models.
\newblock \emph{arXiv preprint arXiv:2407.21783}, 2024.

\bibitem[Exarchos et~al.(2021)Exarchos, Jiang, Yu, and Liu]{exarchos2021policy}
Ioannis Exarchos, Yifeng Jiang, Wenhao Yu, and C~Karen Liu.
\newblock Policy transfer via kinematic domain randomization and adaptation.
\newblock In \emph{2021 IEEE International Conference on Robotics and Automation (ICRA)}, pages 45--51. IEEE, 2021.

\bibitem[Fu et~al.(2024)Fu, Zhao, and Finn]{mobilealoha}
Zipeng Fu, Tony~Z. Zhao, and Chelsea Finn.
\newblock Mobile {ALOHA}: Learning bimanual mobile manipulation using low-cost whole-body teleoperation.
\newblock In \emph{8th Annual Conference on Robot Learning}, 2024.

\bibitem[Geng et~al.(2023)Geng, Li, Geng, Chen, Dong, and Wang]{partmanip}
Haoran Geng, Ziming Li, Yiran Geng, Jiayi Chen, Hao Dong, and He Wang.
\newblock Partmanip: Learning cross-category generalizable part manipulation policy from point cloud observations.
\newblock \emph{arXiv preprint arXiv:2303.16958}, 2023.

\bibitem[Heo et~al.(2023)Heo, Lee, Lee, and Lim]{furniturebench}
Minho Heo, Youngwoon Lee, Doohyun Lee, and Joseph~J. Lim.
\newblock Furniturebench: Reproducible real-world benchmark for long-horizon complex manipulation.
\newblock In \emph{Robotics: Science and Systems}, 2023.

\bibitem[Huber et~al.(2024)Huber, H{\'e}l{\'e}non, Watrelot, Amar, and Doncieux]{huber2024domain}
Johann Huber, Fran{\c{c}}ois H{\'e}l{\'e}non, Hippolyte Watrelot, Fa{\"\i}z~Ben Amar, and St{\'e}phane Doncieux.
\newblock Domain randomization for sim2real transfer of automatically generated grasping datasets.
\newblock In \emph{2024 IEEE International Conference on Robotics and Automation (ICRA)}, pages 4112--4118. IEEE, 2024.

\bibitem[Kerbl et~al.(2023)Kerbl, Kopanas, Leimk{\"u}hler, and Drettakis]{3DGS}
Bernhard Kerbl, Georgios Kopanas, Thomas Leimk{\"u}hler, and George Drettakis.
\newblock 3d gaussian splatting for real-time radiance field rendering.
\newblock \emph{ACM Transactions on Graphics}, 42\penalty0 (4), 2023.

\bibitem[Kumar et~al.(2023)Kumar, Shah, Zhou, Moens, Caggiano, Gupta, and Rajeswaran]{RoboHive}
Vikash Kumar, Rutav Shah, Gaoyue Zhou, Vincent Moens, Vittorio Caggiano, Abhishek Gupta, and Aravind Rajeswaran.
\newblock Robohive: A unified framework for robot learning.
\newblock In \emph{Thirty-seventh Conference on Neural Information Processing Systems Datasets and Benchmarks Track}, 2023.

\bibitem[Li et~al.(2023)Li, Wang, and Tseng]{li2023gaussiandiffusion}
Xinhai Li, Huaibin Wang, and Kuo-Kun Tseng.
\newblock Gaussiandiffusion: 3d gaussian splatting for denoising diffusion probabilistic models with structured noise.
\newblock \emph{arXiv preprint arXiv:2311.11221}, 2023.

\bibitem[Liu et~al.(2024)Liu, Canberk, Song, and Vondrick]{DifferentiableRobot}
Ruoshi Liu, Alper Canberk, Shuran Song, and Carl Vondrick.
\newblock Differentiable robot rendering.
\newblock In \emph{8th Annual Conference on Robot Learning}, 2024.

\bibitem[Long et~al.(2015)Long, Cao, Wang, and Jordan]{long2015learning}
Mingsheng Long, Yue Cao, Jianmin Wang, and Michael Jordan.
\newblock Learning transferable features with deep adaptation networks.
\newblock In \emph{International conference on machine learning}, pages 97--105. PMLR, 2015.

\bibitem[Long et~al.(2023)Long, Guo, Lin, Liu, Dou, Liu, Ma, Zhang, Habermann, Theobalt, et~al.]{wonder3d}
Xiaoxiao Long, Yuan-Chen Guo, Cheng Lin, Yuan Liu, Zhiyang Dou, Lingjie Liu, Yuexin Ma, Song-Hai Zhang, Marc Habermann, Christian Theobalt, et~al.
\newblock Wonder3d: Single image to 3d using cross-domain diffusion.
\newblock \emph{arXiv preprint arXiv:2310.15008}, 2023.

\bibitem[Lou et~al.(2024)Lou, Liu, Pan, Geng, Chen, Ma, Li, Wang, Feng, Shi, et~al.]{lou2024robo}
Haozhe Lou, Yurong Liu, Yike Pan, Yiran Geng, Jianteng Chen, Wenlong Ma, Chenglong Li, Lin Wang, Hengzhen Feng, Lu Shi, et~al.
\newblock Robo-gs: A physics consistent spatial-temporal model for robotic arm with hybrid representation.
\newblock \emph{arXiv preprint arXiv:2408.14873}, 2024.

\bibitem[Lu et~al.(2025)Lu, Zhang, Wang, Liu, Lu, and Tang]{ManiGaussian}
Guanxing Lu, Shiyi Zhang, Ziwei Wang, Changliu Liu, Jiwen Lu, and Yansong Tang.
\newblock Manigaussian: Dynamic gaussian splatting for multi-task robotic manipulation.
\newblock In \emph{European Conference on Computer Vision}, pages 349--366. Springer, 2025.

\bibitem[Makoviychuk et~al.(2021)Makoviychuk, Wawrzyniak, Guo, Lu, Storey, Macklin, Hoeller, Rudin, Allshire, Handa, and State]{isaacgym}
Viktor Makoviychuk, Lukasz Wawrzyniak, Yunrong Guo, Michelle Lu, Kier Storey, Miles Macklin, David Hoeller, Nikita Rudin, Arthur Allshire, Ankur Handa, and Gavriel State.
\newblock Isaac gym: High performance {GPU} based physics simulation for robot learning.
\newblock In \emph{Thirty-fifth Conference on Neural Information Processing Systems Datasets and Benchmarks Track (Round 2)}, 2021.

\bibitem[Matsuki et~al.(2024)Matsuki, Murai, Kelly, and Davison]{GS_SLAM}
Hidenobu Matsuki, Riku Murai, Paul~H.J. Kelly, and Andrew~J. Davison.
\newblock Gaussian splatting slam.
\newblock In \emph{Proceedings of the IEEE/CVF Conference on Computer Vision and Pattern Recognition (CVPR)}, pages 18039--18048, 2024.

\bibitem[Mildenhall et~al.(2021)Mildenhall, Srinivasan, Tancik, Barron, Ramamoorthi, and Ng]{NeRF}
Ben Mildenhall, Pratul~P. Srinivasan, Matthew Tancik, Jonathan~T. Barron, Ravi Ramamoorthi, and Ren Ng.
\newblock Nerf: representing scenes as neural radiance fields for view synthesis.
\newblock \emph{Commun. ACM}, 65\penalty0 (1):\penalty0 99–106, 2021.

\bibitem[Mittal et~al.(2023)Mittal, Yu, Yu, Liu, Rudin, Hoeller, Yuan, Singh, Guo, Mazhar, Mandlekar, Babich, State, Hutter, and Garg]{orbit}
Mayank Mittal, Calvin Yu, Qinxi Yu, Jingzhou Liu, Nikita Rudin, David Hoeller, Jia~Lin Yuan, Ritvik Singh, Yunrong Guo, Hammad Mazhar, Ajay Mandlekar, Buck Babich, Gavriel State, Marco Hutter, and Animesh Garg.
\newblock Orbit: A unified simulation framework for interactive robot learning environments.
\newblock \emph{IEEE Robotics and Automation Letters}, 8\penalty0 (6):\penalty0 3740--3747, 2023.

\bibitem[Mouret and Chatzilygeroudis(2017)]{20years}
Jean-Baptiste Mouret and Konstantinos Chatzilygeroudis.
\newblock 20 years of reality gap: a few thoughts about simulators in evolutionary robotics.
\newblock In \emph{Proceedings of the genetic and evolutionary computation conference companion}, pages 1121--1124, 2017.

\bibitem[{NVIDIA}(2024)]{isaacsim}
{NVIDIA}.
\newblock Isaac sim.
\newblock \url{https://developer.nvidia.com/isaac/sim}, 2024.
\newblock Software.

\bibitem[Qureshi et~al.(2024)Qureshi, Garg, Yandun, Held, Kantor, and Silwal]{SplatSim}
Mohammad~Nomaan Qureshi, Sparsh Garg, Francisco Yandun, David Held, George Kantor, and Abhishesh Silwal.
\newblock Splatsim: Zero-shot sim2real transfer of rgb manipulation policies using gaussian splatting.
\newblock \emph{arXiv preprint arXiv:2409.10161}, 2024.

\bibitem[Radford et~al.(2021)Radford, Kim, Hallacy, Ramesh, Goh, Agarwal, Sastry, Askell, Mishkin, Clark, et~al.]{CLIP}
Alec Radford, Jong~Wook Kim, Chris Hallacy, Aditya Ramesh, Gabriel Goh, Sandhini Agarwal, Girish Sastry, Amanda Askell, Pamela Mishkin, Jack Clark, et~al.
\newblock Learning transferable visual models from natural language supervision.
\newblock In \emph{International conference on machine learning}, pages 8748--8763. PMLR, 2021.

\bibitem[Rombach et~al.(2022)Rombach, Blattmann, Lorenz, Esser, and Ommer]{stablediffusion}
Robin Rombach, Andreas Blattmann, Dominik Lorenz, Patrick Esser, and Bj{\"o}rn Ommer.
\newblock High-resolution image synthesis with latent diffusion models.
\newblock In \emph{Proceedings of the IEEE/CVF conference on computer vision and pattern recognition}, pages 10684--10695, 2022.

\bibitem[Schonberger and Frahm(2016)]{colmap}
Johannes~L Schonberger and Jan-Michael Frahm.
\newblock Structure-from-motion revisited.
\newblock In \emph{Proceedings of the IEEE conference on computer vision and pattern recognition}, pages 4104--4113, 2016.

\bibitem[Shen et~al.(2024)Shen, Cai, Yin, Müller, Li, Wang, Chen, and Wang]{gim}
Xuelun Shen, Zhipeng Cai, Wei Yin, Matthias Müller, Zijun Li, Kaixuan Wang, Xiaozhi Chen, and Cheng Wang.
\newblock Gim: Learning generalizable image matcher from internet videos.
\newblock In \emph{The Twelfth International Conference on Learning Representations}, 2024.

\bibitem[Tobin et~al.(2017)Tobin, Fong, Ray, Schneider, Zaremba, and Abbeel]{tobin2017domain}
Josh Tobin, Rachel Fong, Alex Ray, Jonas Schneider, Wojciech Zaremba, and Pieter Abbeel.
\newblock Domain randomization for transferring deep neural networks from simulation to the real world.
\newblock In \emph{2017 IEEE/RSJ international conference on intelligent robots and systems (IROS)}, pages 23--30. IEEE, 2017.

\bibitem[Todorov et~al.(2012)Todorov, Erez, and Tassa]{mujoco}
Emanuel Todorov, Tom Erez, and Yuval Tassa.
\newblock Mujoco: A physics engine for model-based control.
\newblock In \emph{2012 IEEE/RSJ International Conference on Intelligent Robots and Systems}, pages 5026--5033. IEEE, 2012.

\bibitem[Wang et~al.(2020)Wang, Liu, and Li]{wang2020reinforcement}
Jingkang Wang, Yang Liu, and Bo Li.
\newblock Reinforcement learning with perturbed rewards.
\newblock In \emph{Proceedings of the AAAI conference on artificial intelligence}, pages 6202--6209, 2020.

\bibitem[Whitney et~al.(2018)Whitney, Rosen, Ullman, Phillips, and Tellex]{vr}
David Whitney, Eric Rosen, Daniel Ullman, Elizabeth Phillips, and Stefanie Tellex.
\newblock Ros reality: A virtual reality framework using consumer-grade hardware for ros-enabled robots.
\newblock In \emph{2018 IEEE/RSJ International Conference on Intelligent Robots and Systems (IROS)}, pages 1--9, 2018.

\bibitem[Xiang et~al.(2020)Xiang, Qin, Mo, Xia, Zhu, Liu, Liu, Jiang, Yuan, Wang, Yi, Chang, Guibas, and Su]{sapien}
Fanbo Xiang, Yuzhe Qin, Kaichun Mo, Yikuan Xia, Hao Zhu, Fangchen Liu, Minghua Liu, Hanxiao Jiang, Yifu Yuan, He Wang, Li Yi, Angel~X. Chang, Leonidas~J. Guibas, and Hao Su.
\newblock {SAPIEN}: A simulated part-based interactive environment.
\newblock In \emph{The IEEE Conference on Computer Vision and Pattern Recognition (CVPR)}, 2020.

\bibitem[Zhao et~al.(2023)Zhao, Kumar, Levine, and Finn]{aloha}
Tony~Z. Zhao, Vikash Kumar, Sergey Levine, and Chelsea Finn.
\newblock {Learning Fine-Grained Bimanual Manipulation with Low-Cost Hardware}.
\newblock In \emph{Proceedings of Robotics: Science and Systems}, Daegu, Republic of Korea, 2023.

\bibitem[Zhao et~al.(2020)Zhao, Queralta, and Westerlund]{s2r_survey}
Wenshuai Zhao, Jorge~Peña Queralta, and Tomi Westerlund.
\newblock Sim-to-real transfer in deep reinforcement learning for robotics: a survey.
\newblock In \emph{2020 IEEE Symposium Series on Computational Intelligence (SSCI)}, pages 737--744, 2020.

\bibitem[Zheng et~al.(2023)Zheng, Ma, Cai, Lu, and Wang]{zheng2023gpdan}
Liming Zheng, Wenxuan Ma, Yinghao Cai, Tao Lu, and Shuo Wang.
\newblock Gpdan: Grasp pose domain adaptation network for sim-to-real 6-dof object grasping.
\newblock \emph{IEEE Robotics and Automation Letters}, 8\penalty0 (8):\penalty0 4585--4592, 2023.

\bibitem[Zheng et~al.(2024)Zheng, Chen, Zheng, Gu, Yang, Jin, Li, Zhong, Wang, Liu, Yang, Wang, Chen, Long, and Wang]{GaussianGrasper}
Yuhang Zheng, Xiangyu Chen, Yupeng Zheng, Songen Gu, Runyi Yang, Bu Jin, Pengfei Li, Chengliang Zhong, Zengmao Wang, Lina Liu, Chao Yang, Dawei Wang, Zhen Chen, Xiaoxiao Long, and Meiqing Wang.
\newblock Gaussiangrasper: 3d language gaussian splatting for open-vocabulary robotic grasping.
\newblock \emph{IEEE Robotics and Automation Letters}, 9\penalty0 (9):\penalty0 7827--7834, 2024.

\bibitem[Zhong et~al.(2025)Zhong, Yu, Wu, and Li]{mass3dgs}
Licheng Zhong, Hong-Xing Yu, Jiajun Wu, and Yunzhu Li.
\newblock Reconstruction and simulation of elastic objects with spring-mass 3d gaussians.
\newblock In \emph{European Conference on Computer Vision}, pages 407--423. Springer, 2025.

\end{thebibliography}
